\documentclass[a4paper,conference]{IEEEtran}

\usepackage{booktabs} 
\usepackage{tcolorbox}
\usepackage{tabularx}
\usepackage{array}
\usepackage{colortbl}
\usepackage{amsmath}
\usepackage{caption}
\usepackage{lettrine}
\usepackage{lipsum}
\tcbuselibrary{skins}
\usepackage[hidelinks]{hyperref}

\newcolumntype{Y}{>{\raggedleft\arraybackslash}X}

\tcbset{tab1/.style={fonttitle=\bfseries\large,fontupper=\normalsize\sffamily,
colback=white!10!white,colframe=black!75!black,colbacktitle=Salmon!40!white,
coltitle=black,center title,freelance,frame code={
\foreach \n in {north east,north west,south east,south west}
{\path [fill=black!75!black] (interior.\n) circle (0.5mm); };},}}

\tcbset{tab2/.style={enhanced,fonttitle=\bfseries,fontupper=\normalsize\sffamily,
colback=white!10!white,colframe=black!50!black,colbacktitle=Salmon!40!white,
coltitle=black,center title}}

\title{\textit{SimDeep}: Federated 3D Indoor Localization via Similarity-Aware Aggregation
}

\author{
\begin{tabular}{c c}
    Ahmed Jaheen & Sarah Elsamanody \\
    The American University in Cairo, Egypt & The American University in Cairo, Egypt \\
    ahmed.jaheen@aucegypt.edu & sarahelsamanody@aucegypt.edu \\
    & \\
    Hamada Rizk & Moustafa Youssef \\
    Osaka University, Japan & The American University in Cairo, Egypt \\
    hamada\_rizk@ist.osaka-u.ac.jp & moustafa-youssef@aucegypt.edu \\
\end{tabular}
}

\author{
\IEEEauthorblockN{
Ahmed Jaheen\IEEEauthorrefmark{1},
Sarah Elsamanody\IEEEauthorrefmark{1},
Hamada Rizk\IEEEauthorrefmark{2},
Moustafa Youssef\IEEEauthorrefmark{1}
}
\IEEEauthorblockA{
\IEEEauthorrefmark{1}The American University in Cairo, Egypt\\
Emails: \{ahmed.jaheen, sarahelsamanody, moustafa-youssef\}@aucegypt.edu
}
\IEEEauthorblockA{
\IEEEauthorrefmark{2}Osaka University, Japan, 
Riken-CCS, Japan
Email: hamada\_rizk@ist.osaka-u.ac.jp
}
}

\begin{document}

\maketitle

\begin{abstract}
% Indoor localization is critical for enabling a wide range of location-based services such as navigation, security, and contextual computing in complex indoor environments. Despite significant advances, the deployment of indoor localization systems in real-world settings remains limited due to challenges posed by non-independent and identically distributed (non-IID) data and device heterogeneity. In this paper, we propose \textit{SimDeep}, a novel Federated Learning (FL) framework designed to tackle these challenges. \textit{SimDeep} introduces a Similarity Aggregation Strategy to aggregate model updates based on client similarity, thereby effectively addressing the non-IID issue. Experimental results demonstrate that \textit{SimDeep} achieves 92.89\% accuracy, outperforming traditional federated and centralized techniques, making it a promising solution for practical deployment.

Indoor localization plays a pivotal role in supporting a wide array of location-based services, including navigation, security, and context-aware computing within intricate indoor environments. Despite considerable advancements, deploying indoor localization systems in real-world scenarios remains challenging, largely because of non-independent and identically distributed (non-IID) data and device heterogeneity. In response, we propose \textit{SimDeep}, a novel Federated Learning (FL) framework explicitly crafted to overcome these obstacles and effectively manage device heterogeneity. \textit{SimDeep} incorporates a Similarity Aggregation Strategy, which aggregates client model updates based on data similarity, significantly alleviating the issues posed by non-IID data. Our experimental evaluations indicate that \textit{SimDeep} achieves an impressive accuracy of 92.89\%, surpassing traditional federated and centralized techniques, thus underscoring its viability for real-world deployment. 
\end{abstract}


\section{Introduction}
\lettrine{I}{ndoor} localization has become one of the pivotal modern Location-Based Services, enabling critical applications such as indoor navigation, emergency response, and context-aware services. While Global Positioning Systems (GPS) dominate outdoor positioning, indoor environments poses significant challenges for such systems. That is due to several factors that exist in such environments such as signal degradation and limited satellite visibility. These factors are seen as limitations in complex infrastructures such as multi-floor buildings, where accurate floor-level and room-level localization is essential.

To overcome these challenges, alternative technologies—such as Bluetooth, Ultra-Wideband, inertial sensors, and cellular-based solutions—have been explored \cite{rizk2020ubiquitous,lin2022state,wang2017virtual}.
% \cite{rizk2019solocell} \cite{rizk2018cellindeep} \cite{rizk2015hybrid}
However, these approaches suffer from limitations like spotty coverage, high deployment costs or reliance on specific hardware. In contrast, WiFi-based localization stands out as a scalable and cost-effective solution due to the widespread availability of WiFi infrastructure and WiFi-enabled devices in indoor settings ranging from malls to airports \cite{Abbas2019WiDeep}.

Among WiFi localization techniques, trilateration and RF fingerprinting are the most common. While trilateration estimates user position based on distances from known access points (APs), it suffers from signal variability and environmental interference, especially in vertical (floor-level) localization \cite{chan2012indoor,yang2020novel}. RF fingerprinting offers more precise location estimates by matching real-time signal strengths with a pre-built signal map \cite{dardari2015indoor,wu2019wifi}, but collecting and maintaining this fingerprint database is time-consuming, costly, and often privacy-invasive.

To mitigate privacy concerns, Federated Learning (FL) has emerged as a privacy-preserving solution by enabling devices to collaboratively train models without sharing raw data \cite{li2020federated,ciftler2020federated,yin2020fedloc}. Despite its promise, applying FL to indoor localization faces a core challenge: the highly non-independent and identically distributed (non-IID) nature of signal data across devices and locations.

This non-IID challenge arises from spatial, device-level, and temporal heterogeneity. Devices in different physical environments—such as enclosed offices versus open halls—experience distinct signal distributions. Variability in hardware (e.g., low-end vs. high-end smartphones) introduces discrepancies in signal measurement precision. Additionally, signal characteristics fluctuate over time due to changing human presence, AP interference, and environmental dynamics \cite{hsu2019measuring,li2019convergence,zhao2018federated}. These factors result in imbalanced, unrepresentative, and device-dependent data, making it difficult for a unified global model to generalize across clients.

Existing federated learning methods like FedAvg and FedProx \cite{Li2020Pseudo,gao2022federated} often struggle under such non-IID settings, especially in multi-floor environments where fine-grained floor-level distinctions are critical. Solutions such as personalization and transfer learning \cite{li2020federated,huang2021personalized} have attempted to address the issue, but performance degradation persists, highlighting the need for non-IID-aware aggregation mechanisms to be able to have systems tailored for real-world applications. 

In this paper, we address the non-IID problem by proposing \textit{SimDeep}, an efficient federated learning-based indoor localization system to improve localization-estimation accuracy inspired by \cite{OzekiYRY24}, specifically in multi-floor buildings where the related notion of floor relevance is practically very strong. \textit{SimDeep} begins with a centralized pretraining phase, where an autoencoder learns robust feature representations of received signal strength (RSS). In the federated learning phase, these features are used by multiple clients—representing different devices—to collaboratively train a global model for location prediction, including floor-level estimation, without sharing raw data. The key innovation in SimDeep is the Similarity Aggregation Strategy, a technique designed to evaluate the relevance of client model updates by grouping the clients with their similar characteristics. By aggregating models from these similar clients, the strategy aims to mitigate the effects of non-IID data by approximating an IID setting. This results in a global model that generalizes better across diverse real-world conditions and improves overall localization accuracy.

We further evaluate \textit{SimDeep} on the UJIIndoorLoc dataset, which is a very comprehensive test bed of benchmark data containing multiple buildings and multi-floor entries \cite{Torres-Sospedra2014UJIIndoorLoc}. Our experiments indicate that \textit{SimDeep} substantially outperforms other federated learning methods, such as FedAvg \cite{Li2020Pseudo} and FedProx \cite{gao2022federated}, in dealing with non-IID data and heterogeneous devices with an accuracy of 92.89\%. These findings further illustrate the potential of \textit{SimDeep} as a practical and scalable solution that can be used in many real-world indoor localization applications.

The remainder of this paper is organized as follows. Related work is discussed in section \ref{sec:related-works}. \textit{SimDeep's} system model architecture and the proposed similarity aggregation strategy are outlined in section \ref{sec:model-arch}. The data collection and configuration in addition to the performance evaluation are presented in section \ref{sec:eval}. Finally, the conclusion and future work are summarized in section \ref{sec:conc}.

\section{Related Works}
\label{sec:related-works}
In this section, we focus on the related works for WiFi indoor localization and federated learning techniques.

\subsection{Wi-Fi Fingerprinting Indoor Localization}
Indoor localization has been a focal point of research for many years, with various techniques being explored to enhance accuracy and usability. Among these methods, Wi-Fi fingerprinting has emerged as one of the most significant and widely adopted approaches. Its prominence stems from the ubiquitous presence of Wi-Fi networks and its cost-effective implementation compared to other localization techniques \cite{lin2022state}.

Wi-Fi fingerprinting became extremely based on deep learning techniques as the problem of indoor localization can be perceived as either a classification or a regression problem, depending on the nature of the location data being sought. Classification approaches are employed when the goal is to identify discrete locations, such as determining the building and floor number. In contrast, regression approaches are used to predict continuous coordinates, such as latitude and longitude \cite{gao2022federated}.

Since indoor localization tends to face many issues, various ideas and different models have been proposed to find the most suited approach. For instance many promising models have been proposed based on CNNs \cite{arslantas2024indoor,10472426, ayyalasomayajula2020deep} demonstrating effectiveness in capturing complex spatial patterns in noisy environments. Other approaches utilize RNNs and LSTMs, which are beneficial when Wi-Fi signals vary over time, as shown in \cite{ayinla2024salloc,bai2020dlrnn}. Additionally, autoencoders (AEs) have been used to reduce noise and improve the signal-to-noise ratio in localization tasks, as demonstrated in \cite{Abbas2019WiDeep, 10526268}.

While these models have shown promising results, privacy concerns remain a significant issue across all approaches. The integration of federated learning techniques can address privacy issues by enabling collaborative model training without compromising individual data security. It’s important to note that the choice of model can be adapted based on specific needs and constraints. In fact, many papers did integrate hybrid models to utilize the benefits of different DL disciplines. However, the focus of this paper is not on finding a new model with higher accuracy, but rather on exploring effective solutions concerning FL along with potential real-world implementations.

\subsection{Federated Learning}
Federated Learning (FL) is well known solution for the improvement of data privacy, security issues and communication costs. It facilitates local model training per clients and only applies model weight aggregation; that way keeping client's data hidden \cite{OzekiYRY24}. %Despite that, one constant challenge that keeps facing FL is that it faces a possibility of a much lower accuracy than centralized techniques as training happens in an individual matter. Not to mention, that in the case of dealing with problems that have non-IID data, the aggregation methodologies are not always the appropriate ones % 
However, a persistent challenge for FL is its reduced accuracy compared to centralized methods, particularly under non-IID data distributions \cite{li2020federated}. Thus, FL performance can degrade when data distribution differs among the clients.

A variety of aggregation techniques have been investigated in the domain of research on indoor localization using FL. For example, \cite{Li2020Pseudo} utilized FedAvg, which focuses on aggregating the client models by averaging their weights. Nevertheless, this approach is inadequate in  real-world scenarios where non-IID data is inevitable, leading to a decline in the accuracy and eventually the inability of the model to learn despite its initial performance \cite{ZHU2021371}. Furthermore, FedProx, another aggregation technique that proved to be better in multiple scenarios such as in \cite{gao2022federated}, where it integrates a proximal term to reduce the impact of heterogeneous data \cite{li2020federated}.

As the research in FL techniques progresses, the concept of a personalized federated learning (PFL) has gained significant publicity. That is because PFL adjusts the model aggregation process to account for client-specific characteristics, making it particularly effective in handling non-IID data \cite{wang2019federated,huang2021personalized,dai2022dispfl}. Another notable paper that plays a vital role in our research is \cite{OzekiYRY24}, which introduced an adequate and a practical similarity-based strategy. This technique distinguishes between client models with different data distributions and clusters similar clients together to aggregate their weights and train them together, resulting in more precise predictions even in the presence of data heterogeneity.

Drawing inspiration from the success of \cite{OzekiYRY24} in landslide prediction during its application to similarity aggregation, we decided to investigate the performance of this technique under the framework of a decentralized indoor localization system. We believe that by integrating the similarity aggregation technique, we can reach a solution to the challenges that are concerned with the non-IID data divergence problem and find a way to shift indoor localization research towards tangible, real-world implementation and applications.

\begin{figure}[!tbp]
\vspace{-0.5cm}
  \centering
  \includegraphics[width=1.04\linewidth,keepaspectratio]{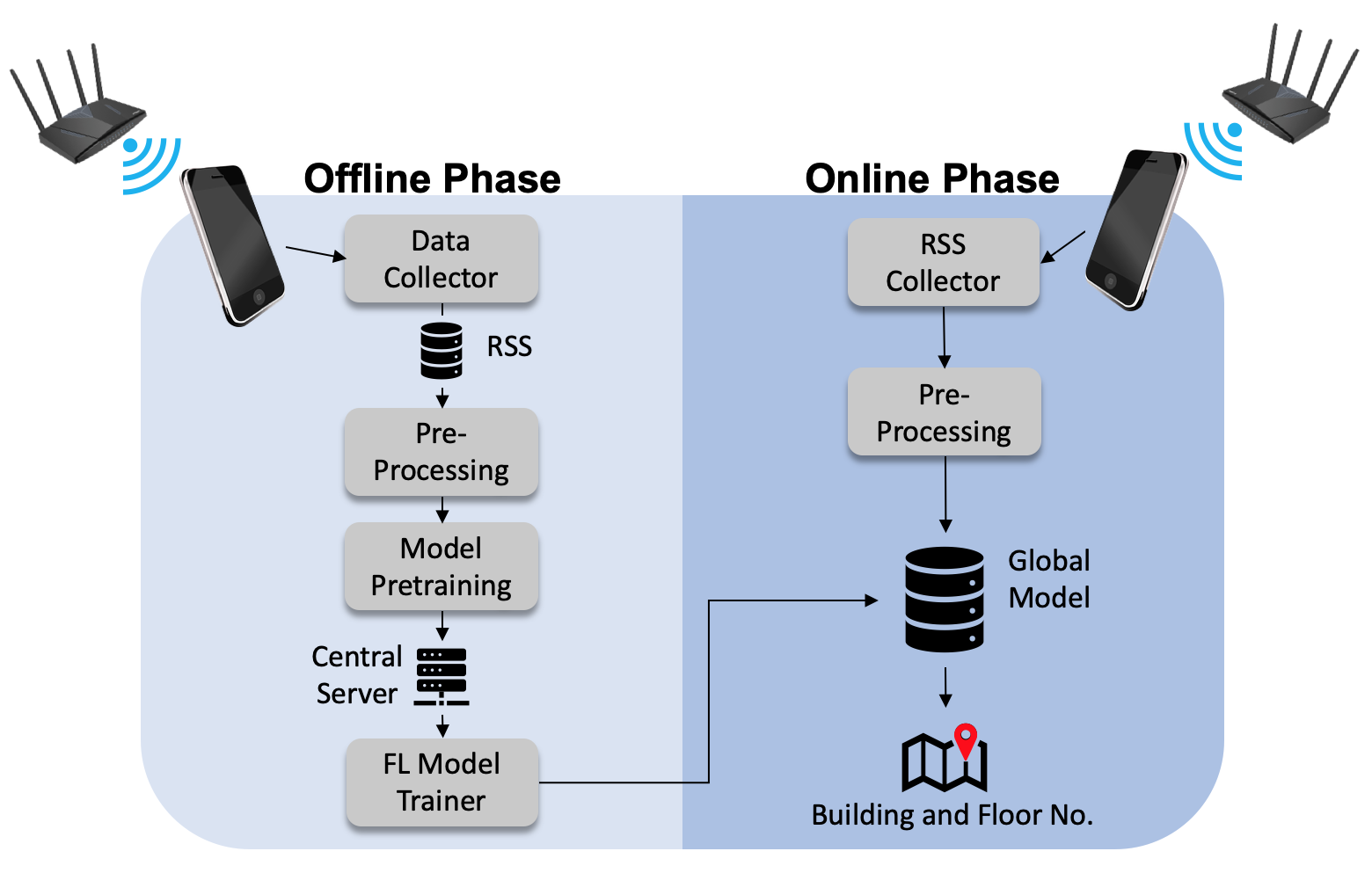}
  \caption{SimDeep System Architecture}
  \label{fig:system}
  \vspace{-0.5cm}
\end{figure}

\section{The SimDeep System}
In this section, we present the details of the different modules of \textit{SimDeep}.
\label{sec:model-arch}

\subsection{Overview}
As illustrated in Figure~\ref{fig:system}, \textit{SimDeep} operates through two main phases: an offline training phase and an online tracking phase.

In the offline training phase, RSS data is collected from Wi-Fi access points (WAPs) within the targeted area. This raw data is first preprocessed to reduce noise and format it appropriately for the model. Each mobile device then uses this preprocessed data to train an initial floor prediction model locally. To further refine these models, a hierarchical federated learning approach is employed. This approach aggregates local models that have similar data distributions, which helps mitigate the effects of non-iid data, thereby improving the overall model accuracy without requiring the sharing of local data. The outcome of this phase is a set of global models, each corresponding to a group of devices with similar data distributions, which are then shared with the respective groups.

During the online tracking phase, the user's device collects current RSS data, which is similarly preprocessed to ensure consistency with the data used during training. This preprocessed data is then fed into the appropriate global model to predict the user's location, specifically identifying the current floor and building. The system also continuously updates its predictions, enabling accurate and seamless real-time tracking of the user's location.
\subsection{Data Preprocessing and Transformation}
Data preprocessing in the \textit{SimDeep} model is a critical stage that prepares raw Wi-Fi fingerprint data for effective model training. The preprocessing module is integrated in both the offline and online phases to ensure consistency at all times. This pipeline incorporates major steps to split the data set into subsets with and without labels, to transform the raw RSSI values and to structure the data into suitable forms for the training of PyTorch-based models.

\subsubsection{Data Splitting}
First, we divide the dataset into a labeled and an unlabeled subset. This is very important for a semi-supervised learning problem, in which one subset of data is to be used along with known labels while the rest are to be utilized to improve the learning capacity of the model without explicit labels \cite{eldele2022label}.

\textbf{Labeled Data Ratio:} In \textit{SimDeep}, some of the data in the dataset are labeled, and some remain unlabeled. This provides the model with a chance to learn from a limited amount of labeled data and then further fine-tuning this prediction from the larger pool of unlabeled data. In this way, it can retain enough data as unlabeled such that pseudo-labeling techniques are explored to help it generalize from the available information \cite{eldele2022label}.

\textbf{Data Subsetting:} Once the labeled data ratio is determined, the dataset is separated into two subsets. The labeled data provides direct supervision, while the unlabeled data enables indirect supervision, allowing the model to benefit from both forms of learning.

\subsubsection{RSSI Value Transformation}

The RSSI values themselves in the dataset are pretty raw. Large variations, with weak signals, will add noise and hence reduces the model's ability to learn useful patterns. In this regard, RSSI values can be transformed into a normalized range that lessens the impact of weak and unreliable signals and puts more emphasis on strong and consistent ones.

\begin{equation}
    \text{new\_rss} =
    \left\{
        \begin{array}{ll}
            0 & \text{if } rss < \min\_rss \text{ or } rss > 0 \\
            \left(\frac{rss - \min\_rss}{-\min\_rss}\right)^{\alpha} & \text{otherwise}
        \end{array}
    \right.
\end{equation}

\textbf{Transformation Process:} The RSSI values are normalized and converted into a value bound to [0,1]. We perform this step by logarithmic scaling, taking into account the minimum RSSI value which is -104 dBm, where $\alpha$ is set to the mathematical exponent e \cite{gao2022federated}. This sorting helps eliminate noise and fluctuations created by weak signals, which in turn improves the quality of input data by focusing only on the strong signals.

\textbf{Feature and Label Extraction:} Further processing is done on the data obtained after its transformation to extract features and labels. The features describe the transformed RSSI values, which represent the signal strengths from the WAPs. The labels are derived using building information and floor information combined, hence providing a comprehensive target for classification, so that the model can guess the class with the highest probability as we are using UJIINDOOR dataset \cite{Torres-Sospedra2014UJIIndoorLoc}. The pre-processing step ensures that the input data is well prepared for training the model along with bringing out the spatial characteristics of the indoor environment clearly.
\begin{figure*}[!tbp]
\centering
% \vspace{-0.5cm}
\begin{minipage}[t]{0.32\textwidth}
    \centering
    \includegraphics[width=\linewidth]{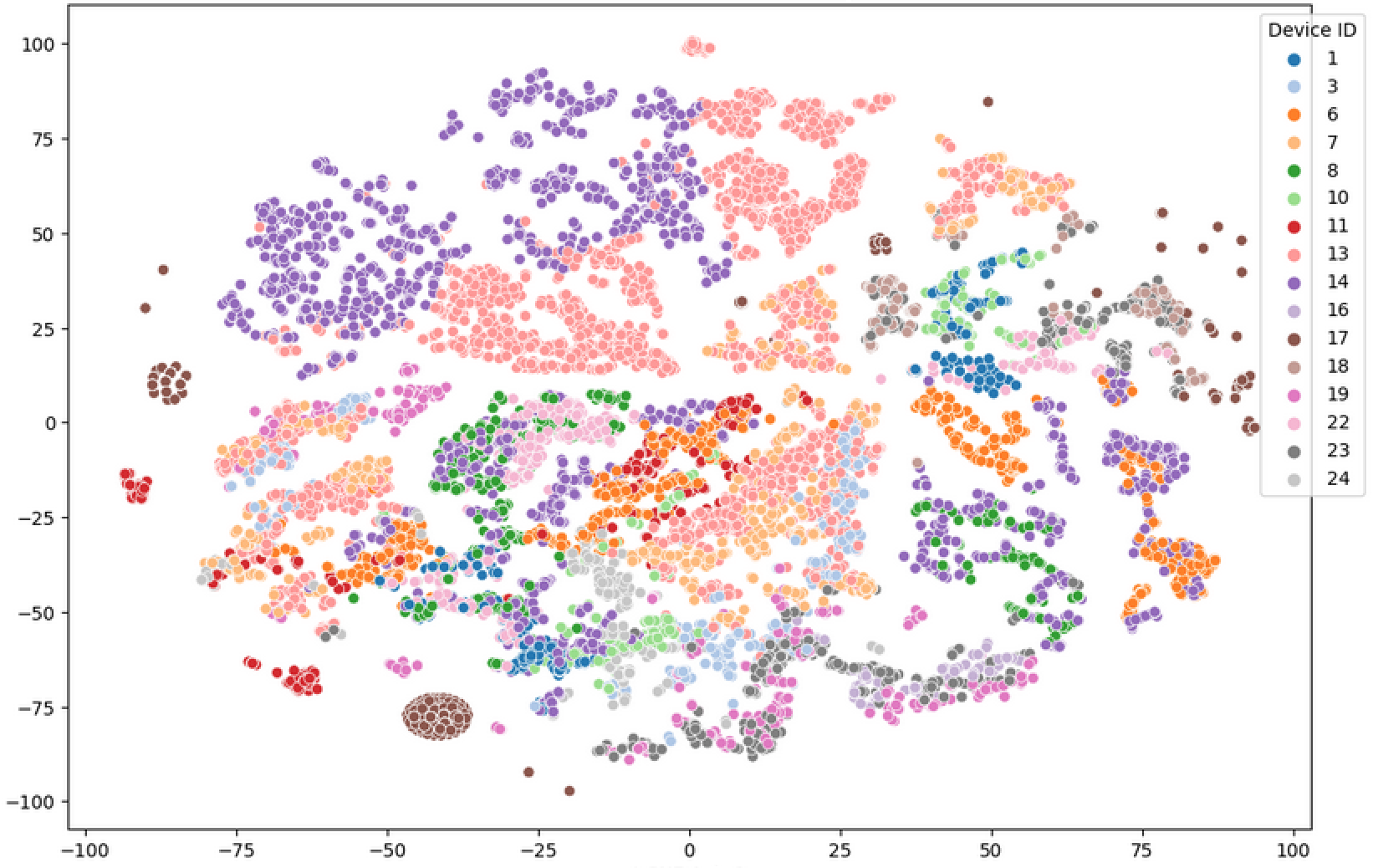}
    \caption{t-SNE projection of raw samples in 2D space for UJIIndoorLoc}
    \label{fig:t-SNE}
\end{minipage}
\hfill
\begin{minipage}[t]{0.32\textwidth}
    \centering
    \includegraphics[width=\linewidth]{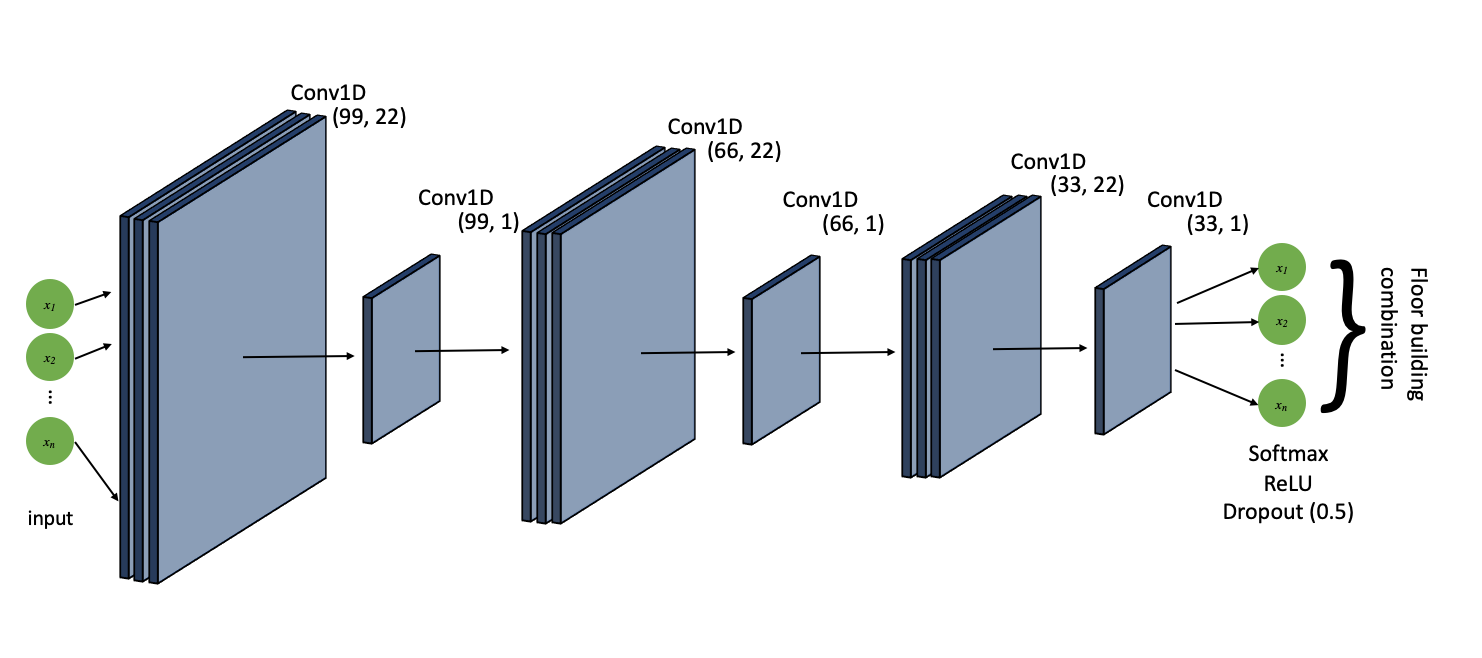}
    \caption{Classifier Architecture}
    \label{fig:classifier}
\end{minipage}
\hfill
\begin{minipage}[t]{0.32\textwidth}
    \centering
    \includegraphics[width=\linewidth]{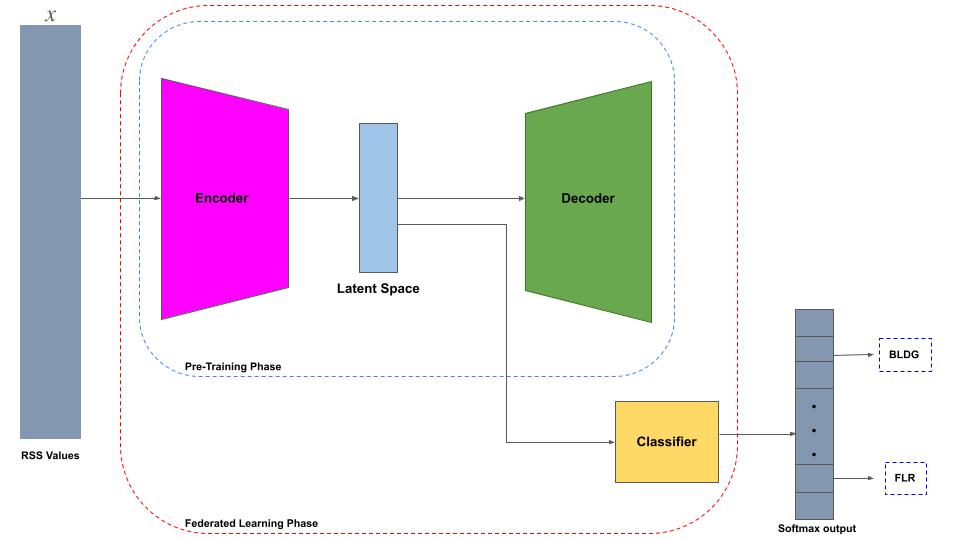}
    \caption{Model Architecture}
    \label{fig:arch}
\end{minipage}
\vspace{-0.5cm}
\end{figure*}

\subsubsection{Creation of PyTorch Datasets \& Non-IID Data}

After feature extraction and labeling, the processed data has to be structured in a specific format that allows for efficient use during model training. For that, the labeled and unlabeled data are structured into datasets compatible with PyTorch. Such datasets enable fast loading of data during training and make sure that the model receives data only in manageable batches. This will also foster easy integration into the structured format with a model's training loop, rendering the processing easy and consistent through various stages of a training process. Also, we use a modified way of partitioning data across clients to simulate non-IID data distributions. Contrasted with the IID data split, in which every client will receive a randomly selected subset of the total dataset, the non-IID strategy will distort this view to obtain a variant distribution better modeling diverse real-world environments. Figure~\ref{fig:t-SNE} highlights how data distribution varies across devices, illustrating the non-IID nature of the problem—where the same data point may be represented differently depending on the device.

\subsection{Model Architecture}
\textit{SimDeep} combines an autoencoder for feature extraction in the pre-training phase and dimensionality reduction with a classifier for predicting the building and floor combination as shown in Figure~\ref{fig:arch}. 

\subsubsection{Autoencoder Architecture}
The autoencoder in \textit{SimDeep} is responsible for extracting the features to be fed into the classifier to enhance the performance. It allows for compressing high dimensionality input features to a much lower and better in application representation. The encoder is structured as three fully connected layers that reduce the dimension from 520 to 64.
% as shown in Figure~\ref{fig:autoencoder}. 
They all follow ReLU activation function and a dropout to regularize it as well as avoid overfitting. The decoder reconstructs the input in a similar manner but we disregard it as we only use the encoder part for the next steps.

% \begin{figure}[!tbp]
%     \centering
%     \includegraphics[width=0.5\textwidth]{Figures/AutoEncoder_Model.pdf}
%     \caption{Autoencoder Architecture}
%     \label{fig:autoencoder}
% \end{figure}

\subsubsection{Classifier for Building-Floor Prediction}
The classifier in \textit{SimDeep} takes the 64-dimensional encoded representation from the autoencoder and predicts the building-floor combination as shown in Figure~\ref{fig:classifier}. The model employs depthwise separable convolutions to efficiently extract features from the encoded input, followed by fully connected layers that output the probability distribution over the possible locations of 15 different classes (combined building and floor). Dropout is further implemented in the classifier to reduce the risk of over-fitting.

With the model architecture outlined, the focus shifts to the pre-training of the feature extraction module (Autoencoder), which enhances the model's ability to handle the data effectively.

% \subsection{Pretraining Phase}
% % After familiarizing with the model architecture, it is important to know that before the federated learning process begins, the model receives centralized pretraining on the server. 
% This step
% is of vital importance as it formulates the foundation of our model. That is because the pretraining allows to strengthens the autoencoder’s ability to learn robust features, leading to a smoother and more effective training process; making the model initialized in a manner that can strongly deal and handle non-IID data that is distributed different among clients.

% \textbf{Pretraining Process:} During this phase, the server trains the autoencoder using a centralized dataset, which represent an aggregated sample of data from various clients. The goal is to learn a pair of functions—an encoder $g(\cdot; \phi)$ and a decoder $h(\cdot; \psi)$—that minimize the reconstruction error between the original input $x_i$ and its reconstruction $\hat{x}_i = h(g(x_i; \phi); \psi)$:

% \begin{equation}
% \min_{\phi, \psi} L_{AE}(\phi, \psi) = \frac{1}{N_0} \sum_{i \in N_0} \|x_i - \hat{x}_i\|^2
% \end{equation}

% This loss function characterizes the difference between the original and reconstructed input data, driving the model to learn an efficient representation of the data.

\subsection{Pre-training of the Feature Extraction module}
This module is crucial for the unsupervised initialization of the feature extraction model. This helps to build a robust and accurate localization model, especially in environments where labeled data is scarce\footnote{Unlabeled data can be easily obtained through crowdsourcing, making it a practical approach for large-scale environments.}. 
This step is designed to prepare the localization model, as detailed in the next section, to effectively manage the non-IID data commonly encountered across different clients.
Each client (mobile device) typically has access to a sufficient amount of unlabeled data along with a smaller subset of labeled data. 

In this process, an autoencoder model is locally trained on each device using its private dataset. The autoencoder, a self-supervised neural network, is specifically employed to learn efficient, latent representations of the input data by using encoder to map it to a compressed latent space from which the decoder network reconstructs it. The key objective here is to utilize the encoder $g(\cdot; \phi)$ from the trained autoencoder as a powerful feature extraction module.

The training process aims to minimize the reconstruction error between the original input $x_i$ and its reconstructed version $\hat{x}_i = h(g(x_i; \phi); \psi)$, with the following loss function:

\begin{equation} \min_{\phi, \psi} L_{AE}(\phi, \psi) = \frac{1}{N_0} \sum_{i \in N_0} |x_i - \hat{x}_i|^2 \end{equation}

This loss function drives the model to learn a compact and meaningful representation of the data, which is less sensitive to noise in the environment or the wireless channels. The encoder, once trained, is used as a feature extraction module that effectively captures the most relevant and intrinsic features of the data.

The resulting autoencoder model, defined by parameters $\phi$ and $\psi$, provides each client with a well-initialized starting point, leading to faster convergence and improved performance in the federated learning phase. By optimizing feature representation locally, this approach lays a strong foundation for creating accurate, personalized models that effectively manage non-IID data, enhancing the performance of decentralized, privacy-preserving localization systems.

After detailing the pre-training process, we turn our attention to the training and federated learning phase, where local training and pseudo-labeling strategies are employed to refine the global model.

\subsection{Local Training and Federated Learning}
\begin{figure*}[!tbp]
\centering
% First Figure
% \vspace{-0.8cm}
\begin{minipage}[t]{0.45\textwidth}
    \centering
    \includegraphics[width=0.85\linewidth, keepaspectratio]{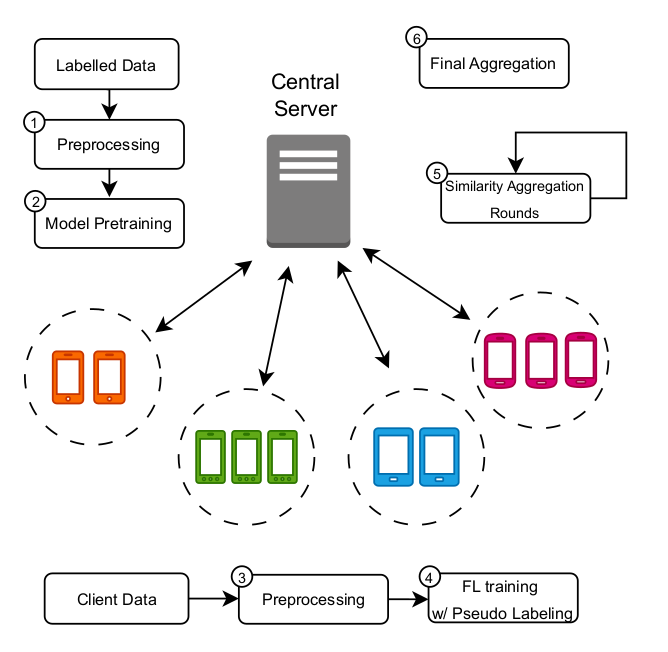}
    \caption{SimDeep Training Flow}
    \label{fig:training}
\end{minipage}
\hfill
% Second Figure
\begin{minipage}[t]{0.45\textwidth}
    \centering
\includegraphics[width=\linewidth]{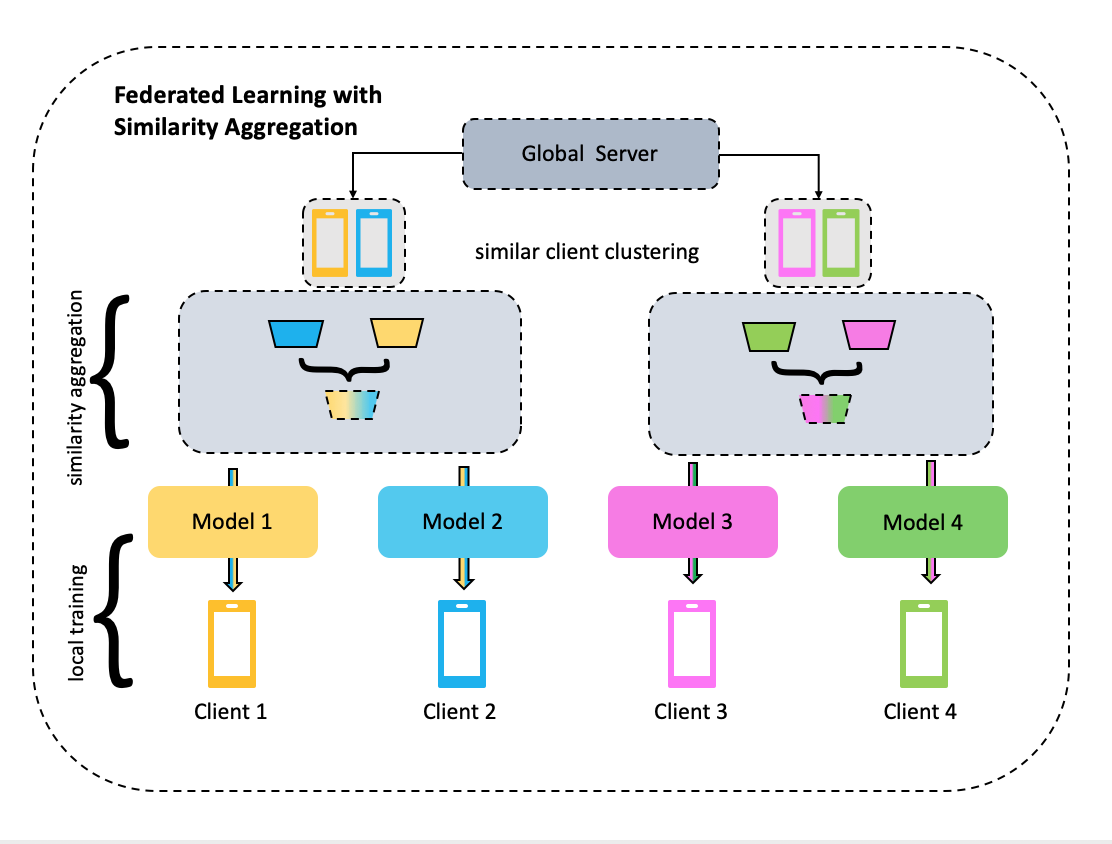}
    \caption{FL Framework with Similarity Aggregation}
    \label{fig:fedratedprocess}
\end{minipage}
\hfill
% Table
\vspace{-0.5cm}
\end{figure*}

Figure~\ref{fig:training} illustrates the training phase of the SimDeep model. This phase encompasses two main processes: local training on each client's dataset and federated learning that refines the global model through the similarity-based aggregation technique as shown in Figure~\ref{fig:fedratedprocess}.

\begin{table}[!tbp]
  \caption{FL Simulation settings.}
  \centering
  \begin{tabular}{l l l}
    \toprule
    \textbf{Parameter} & \textbf{Description} & \textbf{Value} \\
    \midrule
    Optimizer & Model optimizer & Adam \\
    $\eta$ & Learning rate & 0.001 \\
    $\beta_1$, $\beta_2$ & Exponential decay rates & 0.1, 0.999 \\
    C & Number of clients & 10 \\
    B & Batch size & 64 \\
    E & Number of epochs per local iteration & 75 \\
    m & Similarity threshold & 0.5 \\
    R & Communication rounds & 35 \\
    $\gamma$ & Parameter of similarity metrics & 0.5\\ 
    I & Initial rounds & 5 \\
    S & Maximum number of similar clients & 4 \\
    \bottomrule
  \end{tabular}

  \label{tab:fl_simulation_settings}
  \vspace{-0.5cm}
\end{table}
During the federated learning process in \textit{SimDeep}, each client (mobile device) begins by locally train on its private dataset. This local training involves optimizing two parallel loss functions: the reconstruction loss from the autoencoder and the classification loss from the classifier.

\textbf{Autoencoder Loss (MSELoss):} This loss measures the difference between the original input and the reconstructed input of the autoencoder.

\begin{equation}
L_{AE} = \frac{1}{N} \sum_{i=1}^{N} |x_i - \hat{x}_i|^2
\end{equation}

\textbf{Classification Loss (CrossEntropyLoss):} This loss measures the difference between the true labels and the predicted labels.

\begin{equation}
L_{CLS} = -\frac{1}{N} \sum_{i=1}^{N} y_i \log(\hat{y}_i)
\end{equation}

The total loss used for training is a weighted sum of these two losses. The objective remains the same: minimize the reconstruction error of real data distribution at the client. After local training, the server iteratively updates the global model by aggregating the local models:

\begin{align}
\phi^{t+1} &= \frac{1}{\sum_{k \in K} N_k} \sum_{k \in K} N_k \phi_k^{t+1}, \\
\psi^{t+1} &= \frac{1}{\sum_{k \in K} N_k} \sum_{k \in K} N_k \psi_k^{t+1}
\end{align}

where \( N_k \) is the number of data samples in the private dataset of client \( k \). After some iterations, the server achieves the desired global model, denoted by \( (\phi, \psi) \). This process guarantees to have a global model that benefits from diversity in the data across clients while ensuring a unified view and robust feature extraction capability \cite{jamali2022federated,qi2023model}.

Additionally, \textit{SimDeep} leverages pseudo-labeling during the federated learning process to further enhance model performance. In scenarios where unlabeled data is available, pseudo-labeling is applied by having the model assign labels to the unlabeled data based on the highest predicted probability. 
% \cite{lee2013pseudo}. 
This allows the model to utilize additional data effectively, even in the absence of true labels. The pseudo-labels generated from the model’s predictions are then incorporated into the training process, allowing the model to learn from an expanded set of labeled data. Although these pseudo-labels might be noisy, their integration helps in improving predictions and refining the model further, contributing to a more robust federated learning process.

\subsection{Decentralized Collaborative Learning Mechanism}

\textit{SimDeep}’s decentralized collaborative learning mechanism ensures high model performance across clients, even with non-IID distributions. Unlike traditional federated learning, which averages all client model updates, \textit{SimDeep} selects and aggregates updates only from clients with similar data distributions, leading to a more robust and accurate global model.

\textbf{Similarity Calculation:} The similarity between client updates is calculated using the following formula:

\begin{equation}
\text{similarity} = \gamma \cdot \frac{\sum_i (\text{g}_i \cdot \text{g}_j)}{\|\text{g}_i\| \|\text{g}_j\|} + (1 - \gamma) \cdot \frac{\sum_i (\text{acc\_g}_i \cdot \text{acc\_g}_j)}{\|\text{acc\_g}_i\| \|\text{acc\_g}_j\|}
\end{equation} 

Here, $\text{g}_i$ and $\text{g}_j$ represent the gradients of the weights from client $i$ and $j$, and $\text{acc\_g}_i$ and $\text{acc\_g}_j$ are the accumulated gradients. The parameter $\gamma$ controls the balance between the instantaneous gradient similarity and the accumulated gradient similarity, ensuring that both recent updates and historical trends contribute to the similarity calculation \cite{OzekiYRY24}.

\textbf{Selective Similarity-Based Aggregation Process:} To optimize the aggregation of client updates, \textit{SimDeep} employs a selective strategy where only the updates from the most similar clients are aggregated. This method ensures convergence and minimizes computational complexity, making the aggregation process both efficient and effective.

In \textit{SimDeep}, each client $i$ locally updates its model parameters $w_i$ over multiple epochs before transmitting these updates to the global server. The global server then calculates the similarity between updates from different clients using a predefined metric. For each client $i$, the server identifies a set of \textbf{neighbor clients} whose similarity scores exceed a specified threshold. These neighbors are selected based on the similarity of their model updates, ensuring that only relevant and closely aligned data distributions are aggregated.

\textbf{Model Update:} The selected updates from these similar clients are then aggregated to form a new global model :

\begin{equation}
w_i^{t+1} = \frac{1}{|S_i|} \sum_{j \in S_i} w_j^{t+1}
\end{equation}

where $S_i$ is the set of similar clients to client $i$, and $w_j^{t+1}$ represents the model parameters of client $j$ at round $t+1$ \cite{OzekiYRY24}.

With the decentralized collaborative learning mechanism established, we will now evaluate the performance of \textit{SimDeep}'s global model to assess the effectiveness of the proposed strategies in achieving robust and accurate predictions

\section{Evaluation}
\label{sec:eval}
In this section, we compare the performance of \textit{SimDeep} against other federated learning methods, such as FedAvg and FedProx, and against centralized learning on the UJIIndoorLoc dataset. We begin through analysis of the effect of various parameters toward performance, and look at the robustness of the system under various scenarios.

\subsection{Data Collection and Configuration}
We divided the UJIIndoorLoc dataset into two sub datasets: one for training (19,937 RSSI recordings) and one for validation (1,111 recordings), captured four months apart to simulate real-world conditions \cite{Torres-Sospedra2014UJIIndoorLoc}. To make a fair comparison of the results with state-of-the-art methods, we have divided the training data with a ratio of 70:30 for model training and testing. The main goal of \textit{SimDeep} is to get the best classification accuracy over a non-IID setup while preserving privacy in FL. Additionally, the validation dataset is used to measure the performance metrics of our model for a benchmark analysis with the other methods under non-IID conditions. Another line of discussion would be how similarity thresholds, the number of clients and maximum similar clients impact accuracy. We chose FedAvg and FedProx as baselines due to their popularity and foundational role in federated learning. Both are widely adopted in FL benchmarking, especially under non-IID settings, making them a fair and informative comparison point for evaluation. We adopted a semi-supervised learning setup, where 30\% of the training data is labeled and the remaining 70\% is treated as unlabeled. Pseudo-labels are refreshed every 5 communication rounds using a confidence threshold of 0.8, only predictions with high certainty are used for training to mitigate label noise and improve generalization. All experiments were conducted using Kaggle notebooks with a T4 GPU, 16 GB RAM, and 12-hour session limits. In addition, our final FL system configuration is shown in Table~\ref{tab:fl_simulation_settings}. 

%We are working on kaggle as a machine environment due to the limited resources; hence, we do not have the luxury of evaluating different parameters with different tests.%
%Lastly, you can find our GitHub repository and access it through here: \href{https://github.com/mego74/SimDeep}{Project Repository}.

\subsection{SimDeep Evaluation}

\subsubsection{Building and Floor Classification Accuracy}
\begin{figure*}[!tbp]
  \centering
    % \vspace{-0.8cm}
  \begin{minipage}[b]{0.32\textwidth}
    \centering
\includegraphics[width=\textwidth, height=0.7\textwidth, keepaspectratio]{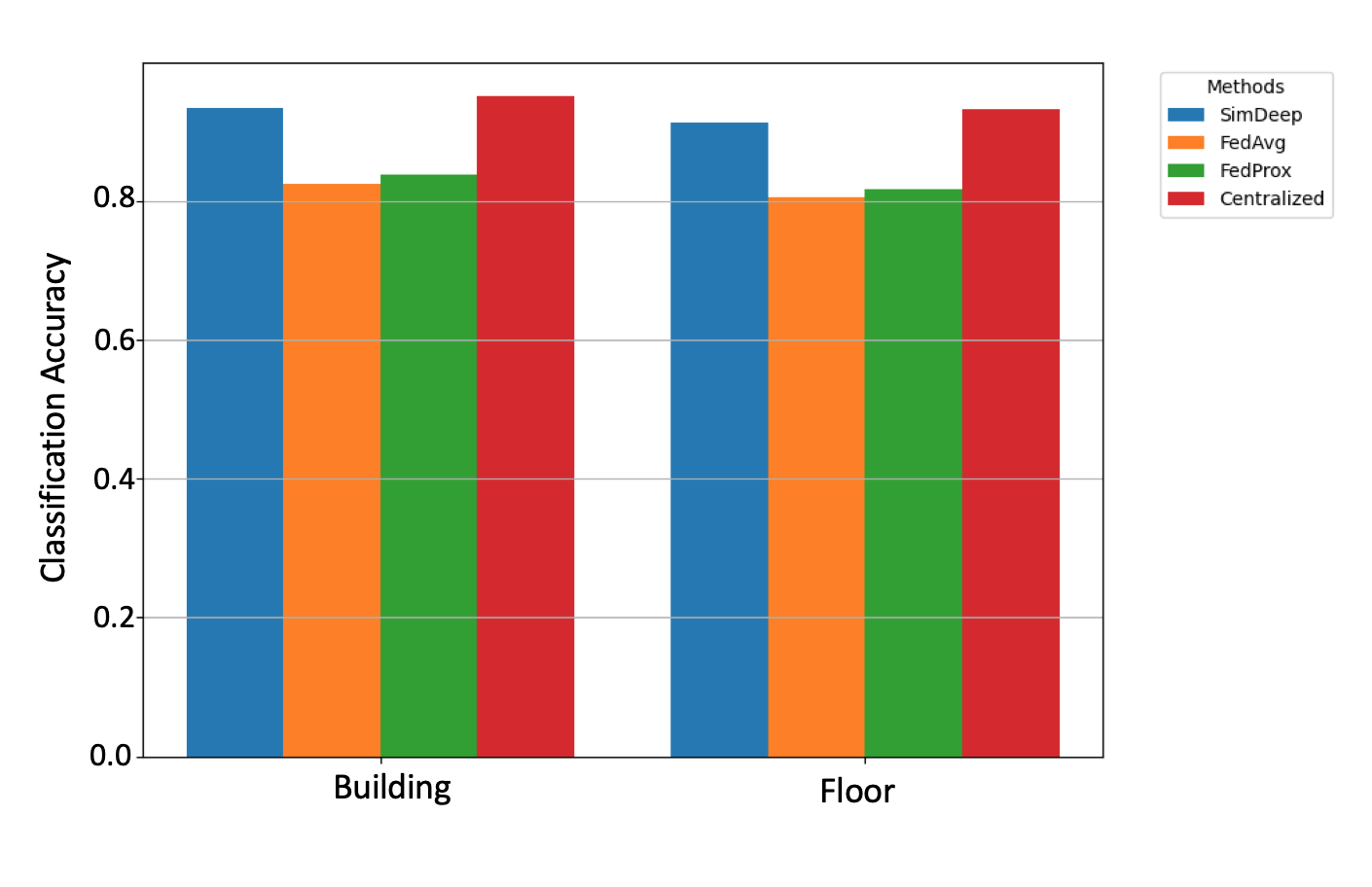}
    \caption{Building and floor classification accuracy for different methods}
\label{fig:building_vs_floor_accuracy}
  \end{minipage}
  \begin{minipage}[b]{0.32\textwidth}
    \centering
\includegraphics[width=\textwidth, height=0.6\textwidth, keepaspectratio]{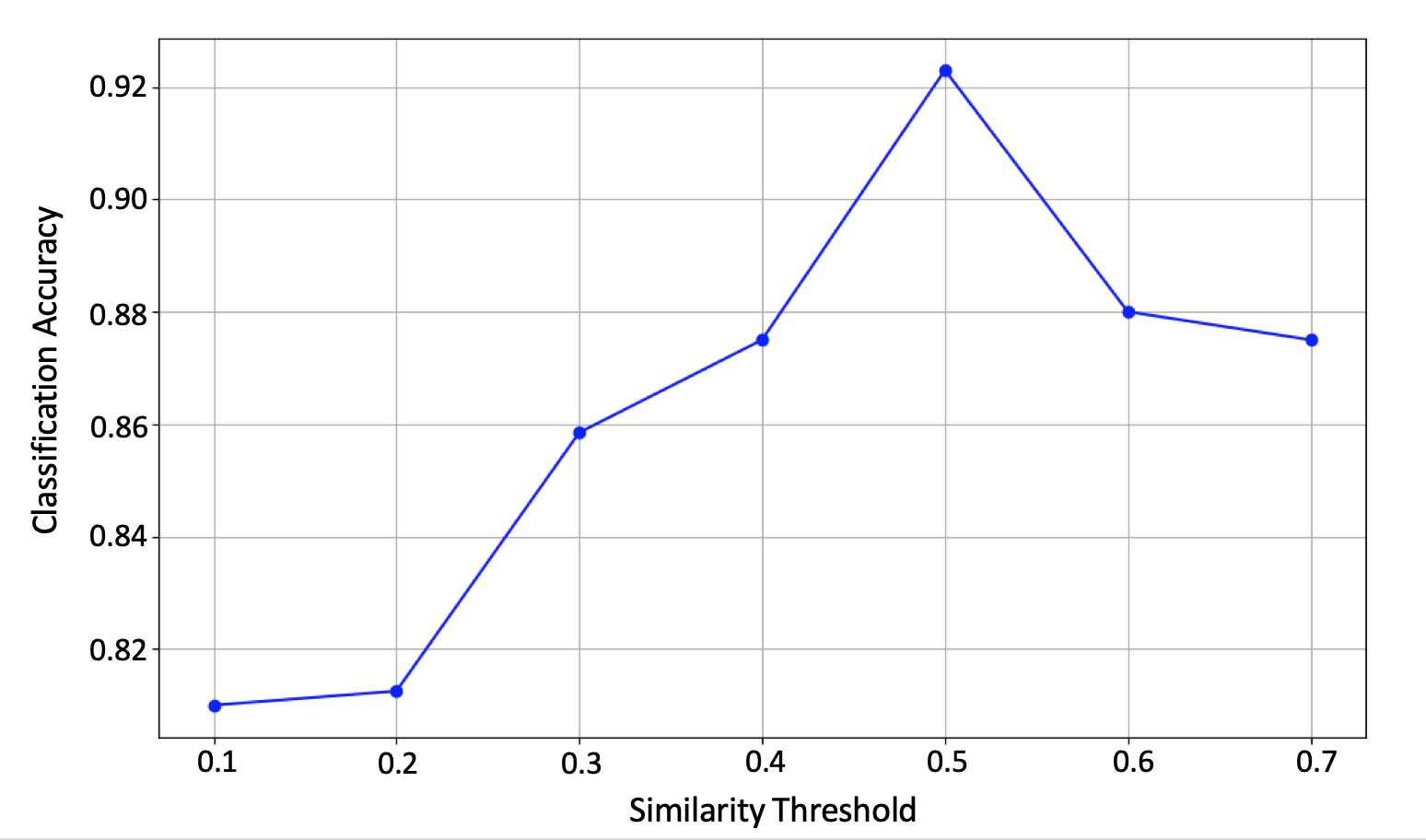}
    \caption{Impact of similarity threshold on classification accuracy.}
\label{fig:threshold_impact}
  \end{minipage}
  \begin{minipage}[b]{0.32\textwidth}
    \centering
\includegraphics[width=\textwidth, height=0.6\textwidth, keepaspectratio]{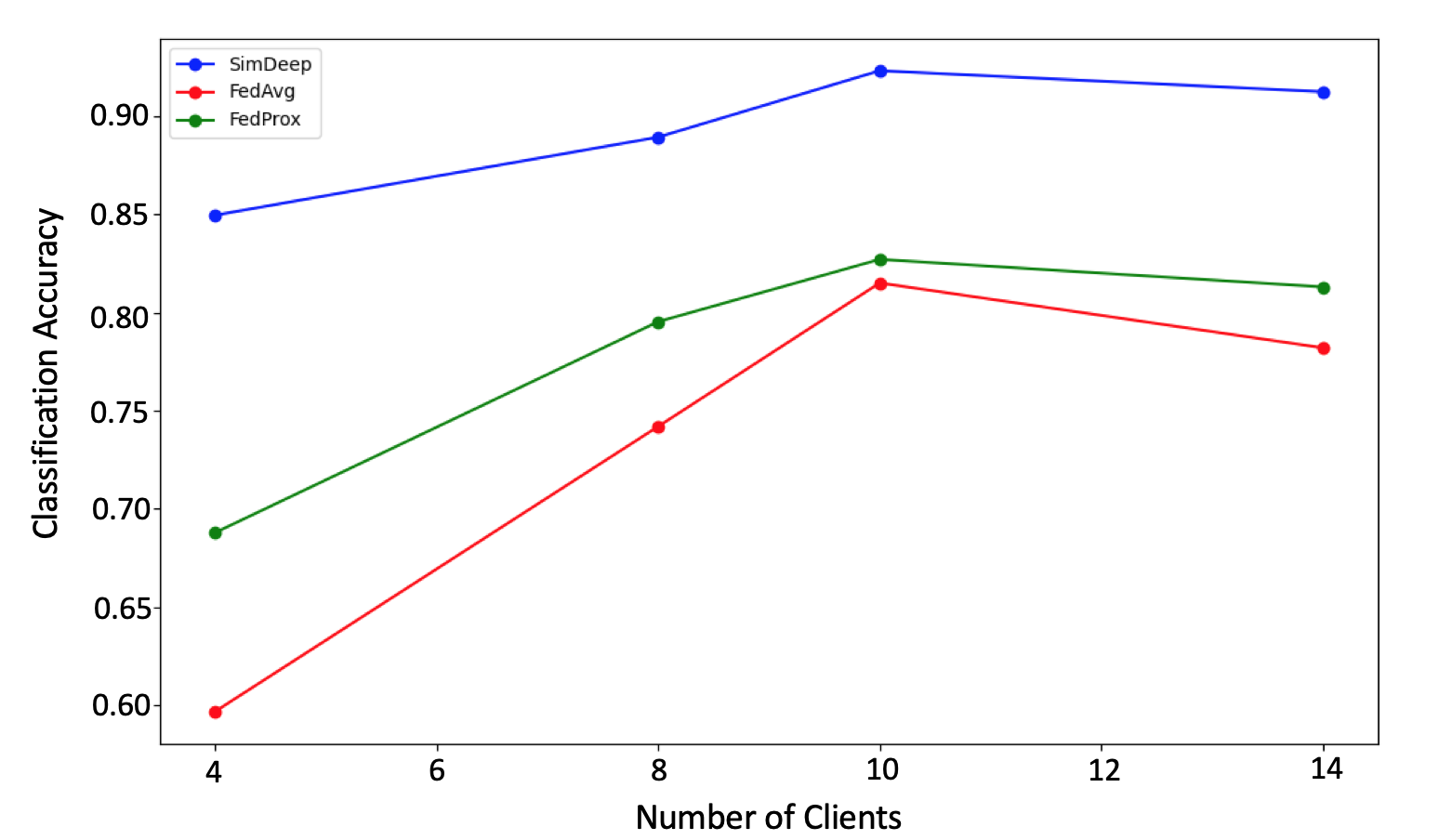}
    \caption{Impact of the number of clients on classification accuracy.}
\label{fig:clients_impact}
  \end{minipage}
  \hfill
  \begin{minipage}[b]{0.32\textwidth}
    \centering
\includegraphics[width=\textwidth, height=0.6\textwidth, keepaspectratio]{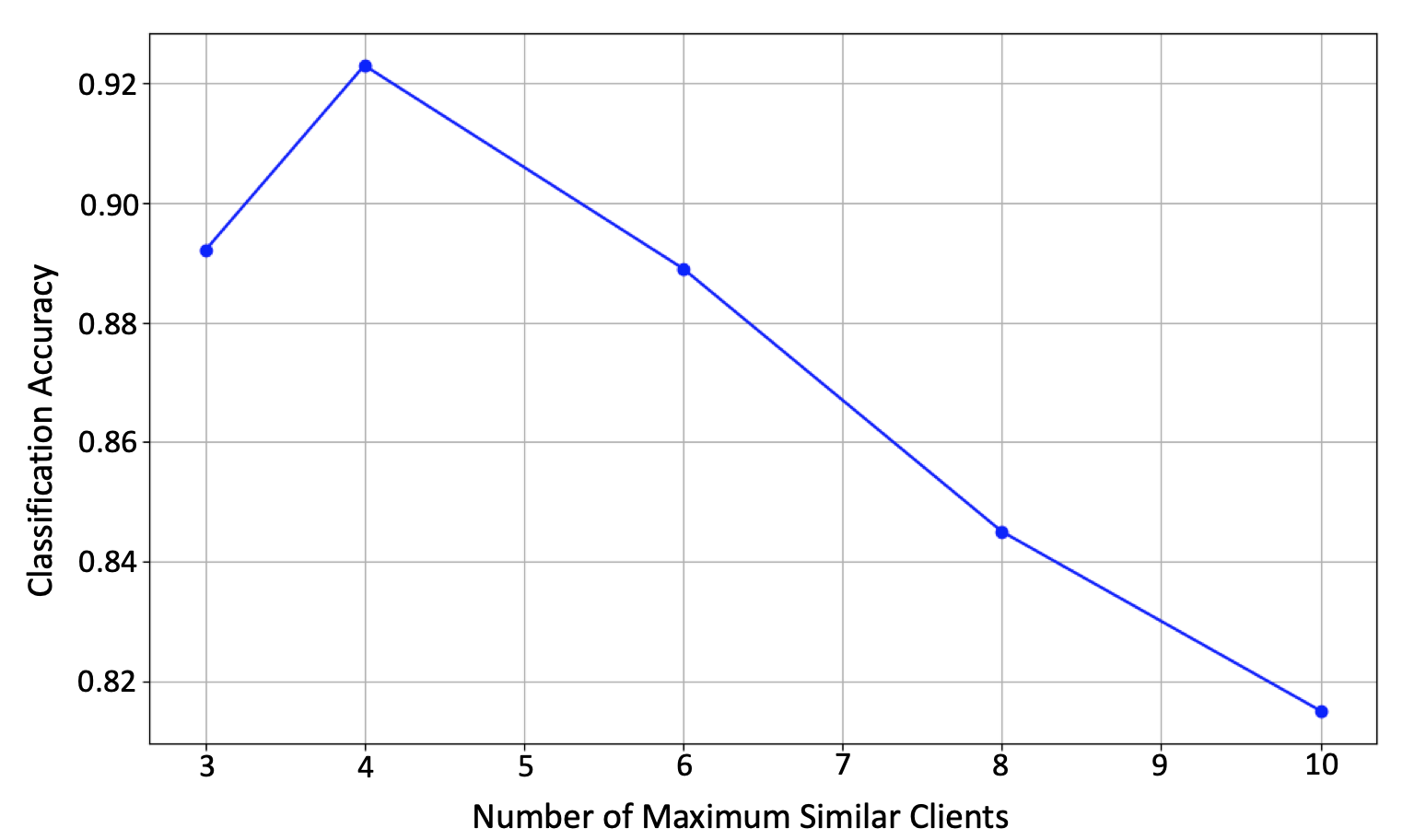}
    \caption{Impact of the maximum number of similar clients on classification accuracy.}
\label{fig:similar_clients_impact}
  \end{minipage}
  \begin{minipage}[b]{0.32\textwidth}
    \centering
    \includegraphics[width=\textwidth, height=0.7\textwidth, keepaspectratio]{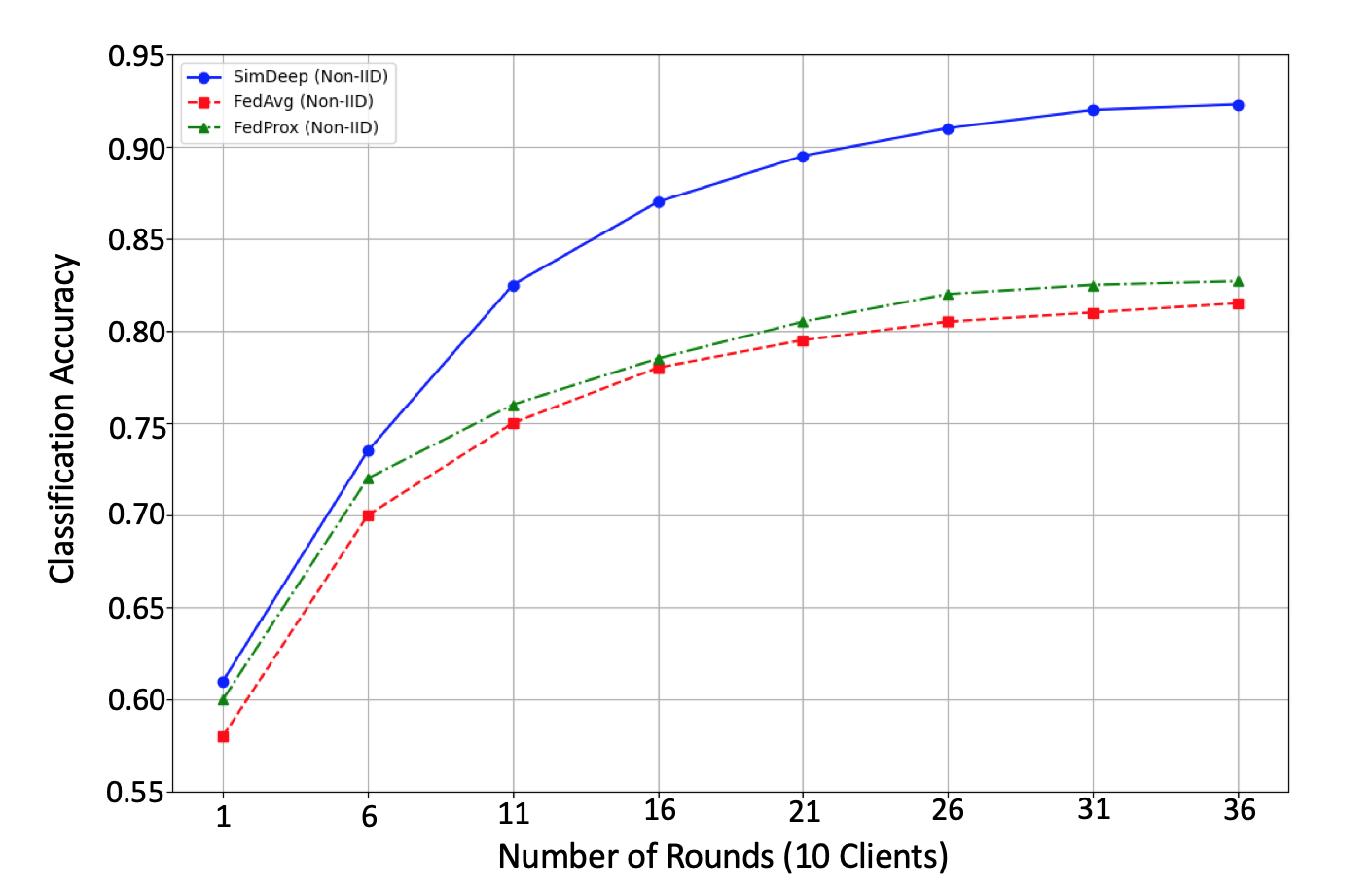}
    \caption{Impact of the number of communication rounds on accuracy.}
    \label{fig:rounds_impact}
  \end{minipage}
  \hfill
  \begin{minipage}[b]{0.32\textwidth}
    \centering
    \includegraphics[width=\textwidth, height=0.7\textwidth, keepaspectratio]{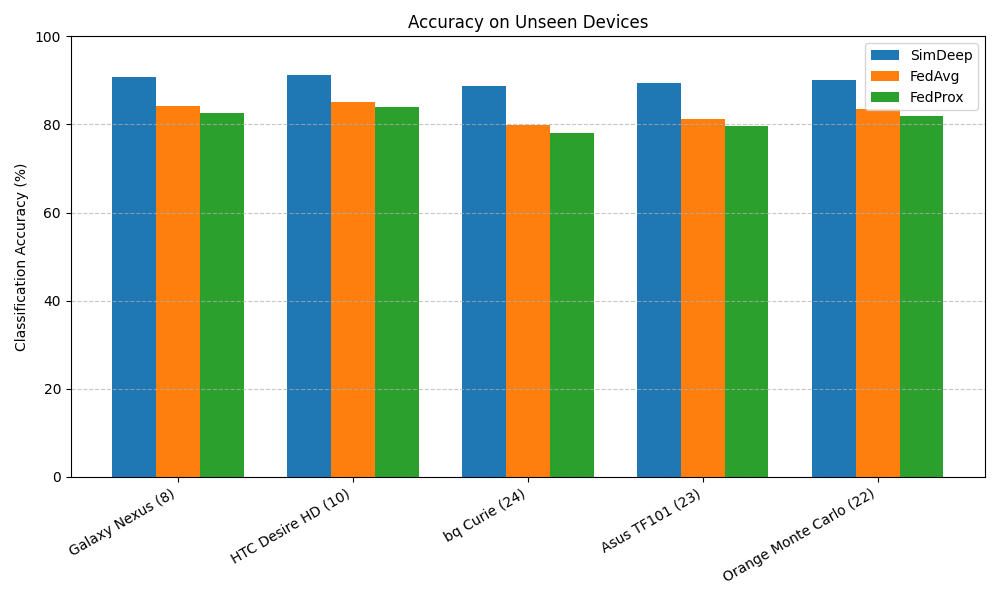}
    \caption{Classification accuracy on unseen test devices.}
    \label{fig:device_heterogeneity}
  \end{minipage}
  \vspace{-0.5cm}
\end{figure*}

The dataset contains metrics for classification accuracy in building and floor recognition. The results depict the efficiency of each FL approach. Figure~\ref{fig:building_vs_floor_accuracy} plots a comparison of the accuracy for building and floor classification tasks across methods, knowing that class 0 means Building 1 Floor 1, class 1 means Building 1 Floor 2, class 5 means Building 2 Floor 1, and so on. \textit{SimDeep} has higher accuracies for building and floor classification compared to FedAvg and FedProx. Notably, \textit{SimDeep} maintains accuracy levels comparable to those achieved through centralized training, while still preserving client privacy. This improvement may be attributed to \textit{SimDeep}'s ability to handle the specific features of the data more effectively, especially in non-IID environments, where FedAvg and FedProx may struggle due to their varying requirements \cite{baumgart2024federatedlearningalgorithmscreated}.

\subsubsection{Impact of Similarity Threshold}
The threshold of similarity $m$ is a parameter that ranges between 0.1 and 0.7, which impacts directly on the process of client grouping. We assume that a threshold of 0.5 offers a balanced trade-off between the similarity of clients and the full model knowledge aggregation. Similarity thresholds are plotted with respect to classification accuracy for \textit{SimDeep} in Figure~\ref{fig:threshold_impact}. For \textit{SimDeep}, the highest accuracy occurs at a threshold level of 0.5. It should be attributed to the fact that at this threshold, clients are close enough to be similar and benefit from knowledge sharing, yet still diverse enough to avoid overfitting. Our findings align with those in \cite{OzekiYRY24}. A lower threshold limits knowledge breadth by aggregating only highly similar clients and overfitting, while a higher threshold introduces noise and weakens collaboration.

%This is likely because clients are close enough to be similar and benefit from knowledge sharing, yet still diverse enough to avoid overfitting. Our finding also aligns with that found in \cite{OzekiYRY24}. When the threshold is too low, it overfits to overly similar clients; too high, and the aggregation becomes noisy and ineffective.

\subsubsection{Impact of Number of Clients}

The number of clients ranged from 2 to 10 to show the impact on model accuracy and scalability. Figure~\ref{fig:clients_impact} illustrates the impact of the number of clients over classification accuracy. By varying the number of clients, \textit{SimDeep} always performs better compared to FedAvg and FedProx. In particular, it attains the best accuracy where the latter two methods have 10 clients. In general, the performance improvement by increasing the number of clients can be explained by the fact that the proposed \textit{SimDeep} aggregator trains more robust and accurate models effectively aggregating very diverse data from multiple clients, but it starts to maintain relatively the same accuracy after 10 clients. It also compares favorably to FedAvg and FedProx, which, while improved by increased clients, do not achieve this same level of accuracy, likely due to limitations in their model aggregation and client collaboration mechanisms \cite{baumgart2024federatedlearningalgorithmscreated}.

%What is this?
\begin{figure}[!tbp]
  \centering
  
\end{figure}

\subsubsection{Impact of Maximum Number of Similar Clients}
It is noticed that the number of similar clients, which is the number of clients having a high degree of similarity in data, significantly impacts model performance. For this, we have changed the number of similar clients from 3 to 10 and checked the classification accuracy of \textit{SimDeep}. Figure~\ref{fig:similar_clients_impact} shows the effect when it has as many as 10 similar clients in classification accuracy. \textit{SimDeep} achieves peak performance when aggregating updates from 4 similar clients. This result was determined empirically through trial and error, where grouping more than 4 clients led to a decline in accuracy. This drop is likely caused by increased divergence in model updates, emphasizing the advantage of aggregating from a moderately sized group with high similarity.

%It, however, started falling at an accuracy of around when the number of clients with similar characteristics went up to 10, as it is then that the model begins to diverge. This optimal performance underscores the benefit of having a group of moderate size with similar clients since more effective aggregation of knowledge and model fine-tuning are allowed.

\subsubsection{Impact of Number of Rounds}
The number of communication rounds is a key factor in federated learning convergence and final accuracy. We ran \textit{SimDeep}, FedAvg, and FedProx for various numbers of communication rounds to evaluate their performance. Figure~\ref{fig:rounds_impact} illustrates the classification accuracy as a function of the number of communication rounds. For all methods, an increase in the number of rounds generally implies better convergence and refinement of the model. The accuracy of \textit{SimDeep} improves significantly, reaching its peak after 35 rounds. While FedAvg and FedProx also show improved accuracy with additional rounds, their performance peaks at a lower accuracy compared to \textit{SimDeep}.

\subsubsection{Performance Under Device Heterogeneity}
Although the training dataset includes samples from 16 different mobile devices, we isolate five specific devices that were not used in validation during training (PhoneIDs 8, 10, 22, 23, and 24) to be able to test properly for the device heterogeneity consistency. Figure~\ref{fig:device_heterogeneity} shows the classification accuracies of \textit{SimDeep}, FedAvg, and FedProx on these unseen devices, testing for the same locations. \textit{SimDeep} achieves consistent and higher accuracy across all devices, ranging from 88.7\% to 91.2\%, whereas FedAvg and FedProx show substantial drops in performance, particularly on low-end or less common devices. These results confirm the robustness of \textit{SimDeep}'s performance to generalize across varying signal distributions introduced by hardware differences, confirming its resilience to device-level non-IID conditions.

\subsubsection{Final Accuracy Performance}
\begin{figure}[!t]
  \centering
  % \vspace{-0.8cm}
  \begin{minipage}[b]{0.52\textwidth}
    \centering
    \includegraphics[width=\textwidth, height=0.7\textwidth, keepaspectratio]{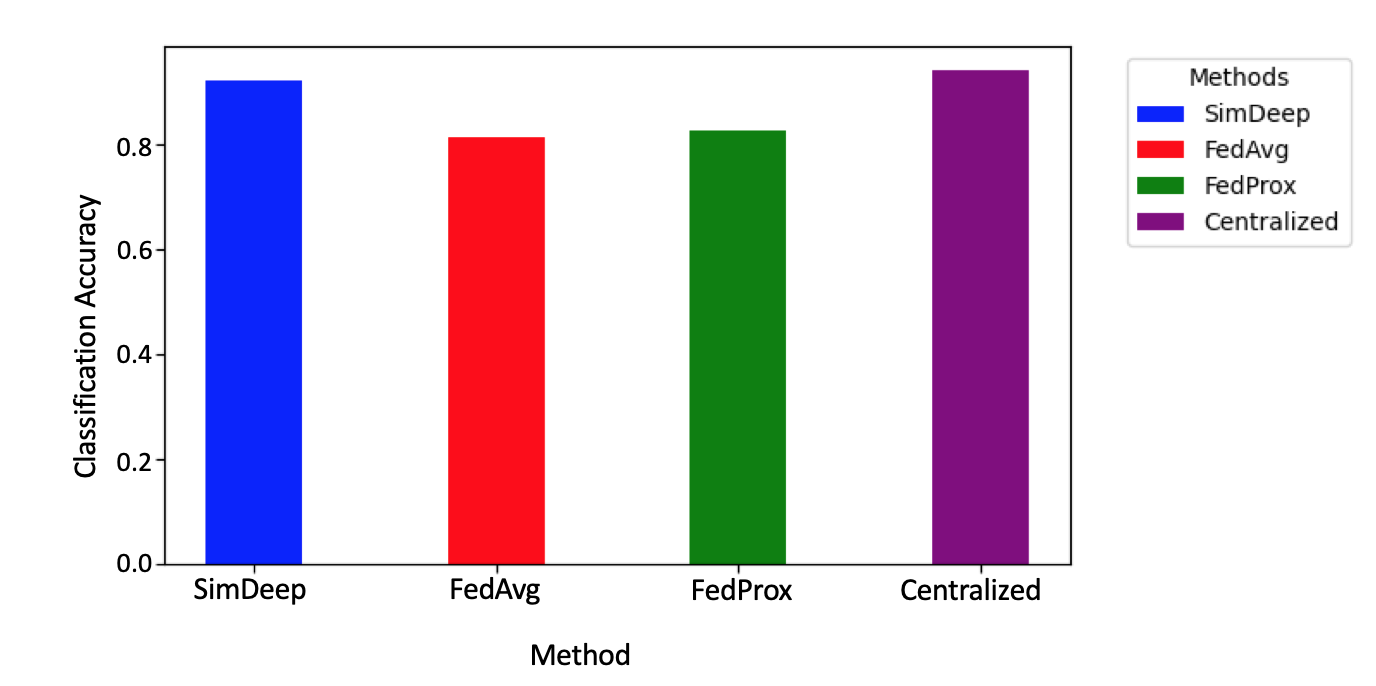}
    \vspace{-0.7cm}
    \caption{Comparison of final accuracy metrics.}
\label{fig:accuracy_comparison}
  \end{minipage}
  \vspace{-1cm}
\end{figure}

Figure~\ref{fig:accuracy_comparison} compares final accuracy metrics for \textit{SimDeep}, FedAvg, and FedProx. It is obvious that \textit{SimDeep} demonstrates the highest level of accuracy compared with centralized learning, while FedAvg and FedProx have lower levels of accuracy. Probably the reason behind the resilience of \textit{SimDeep} to non-IID data lies in the sophisticated way it handles client diversity and knowledge aggregation; the algorithm becomes resilient against heterogeneity in data distributions.

%\subsection{Limitations} While \textit{SimDeep} demonstrates strong performance and robustness to non-IID data, there are limitations to consider. First, the similarity aggregation strategy, while effective, just like any aggregation method it still introduces computational overhead at the server when calculating similarity scores between clients. Additionally, the use of a fixed similarity threshold may not adapt optimally across different deployment environments. Another limitation lies in the reliance on pseudo-labeling, which may introduce noise when confidence scores are miscalibrated, especially in edge cases. Finally, while the UJIIndoorLoc dataset offers a robust benchmark, future validation across more diverse environments (e.g., hospitals, airports) and real-world deployments would further strengthen the generalizability of SimDeep.

\section{Conclusion}
\label{sec:conc}
This paper introduces \textit{SimDeep}, a federated learning approach that advances beyond traditional methods for handling non-IID data. By employing a unique architecture with similarity-based mechanisms that leverage collaborative learning and a semi-supervised training pipeline, \textit{SimDeep} improves building and floor classification accuracy while preserving data privacy. Evaluated on the UJIIndoorLoc dataset, it achieves 92.89\% accuracy, outperforming classical FL methods such as FedAvg \cite{Li2020Pseudo} and FedProx \cite{gao2022federated} by over 8\%. While effective, \textit{SimDeep} introduces server-side computational overhead from pairwise similarity calculations. It also relies on fixed thresholds and uses pseudo-labeling—which, though useful, may affect adaptability if miscalibrated. In future work, we plan to refine the architecture, perform large-scale experiments on higher-performance hardware, build a custom dataset for deployment testing in our university as a real-world environment, and explore its applications in other challenging domains.

\bibliographystyle{IEEEtran}
\bibliography{_ref_SimDeep}

\end{document}